\pdfoutput=1

\documentclass[11pt]{article}

\usepackage[final]{acl}

\usepackage{times}
\usepackage{latexsym}

\usepackage[T1]{fontenc}

\usepackage[utf8]{inputenc}

\usepackage{microtype}

\usepackage{inconsolata}

\usepackage{graphicx}

\usepackage{booktabs}
\usepackage{multirow}
\usepackage{tabularx}
\usepackage{amssymb}
\usepackage{amsmath} 
\usepackage{makecell}
\usepackage{ulem}
\usepackage{subfigure}

\title{Enhancing Tool Retrieval with Iterative Feedback from \\ Large Language Models}

\author{Qiancheng Xu, Yongqi Li$^{\dagger}$, Heming Xia, Wenjie Li \\
Department of Computing, The Hong Kong Polytechnic University, China \\
\texttt{\{qiancheng.xu, he-ming.xia\}@connect.polyu.hk} \\
\texttt{liyongqi0@gmail.com} \quad
\texttt{cswjli@comp.polyu.edu.hk}
}

\begin{document}
\maketitle
\begingroup\def\thefootnote{$\dagger$}\footnotetext{Corresponding author.}\endgroup
\begin{abstract}
Tool learning aims to enhance and expand large language models' (LLMs) capabilities with external tools, which has gained significant attention recently. Current methods have shown that LLMs can effectively handle a certain amount of tools through in-context learning or fine-tuning. However, in real-world scenarios, the number of tools is typically extensive and irregularly updated, emphasizing the necessity for a dedicated tool retrieval component. Tool retrieval is nontrivial due to the following challenges: 1) complex user instructions and tool descriptions; 2) misalignment between tool retrieval and tool usage models. To address the above issues, we propose to enhance tool retrieval with iterative feedback from the large language model. Specifically, we prompt the tool usage model, i.e., the LLM, to provide feedback for the tool retriever model in multi-round, which could progressively improve the tool retriever's understanding of instructions and tools and reduce the gap between the two standalone components. We build a unified and comprehensive benchmark to evaluate tool retrieval models. The extensive experiments indicate that our proposed approach achieves advanced performance in both in-domain evaluation and out-of-domain evaluation\footnote{Code available at 
\href{https://github.com/travis-xu/TR-Feedback}{https://github.com/travis-xu/TR-Feedback}.}.
\end{abstract}

\section{Introduction}
Large language models (LLMs) have demonstrated remarkable success in language-related tasks and are considered a potential pathway to achieving artificial general intelligence~\cite{zhao2023survey}. 
However, despite their powerful capabilities, LLMs are still limited in many aspects, such as knowledge update and mathematical reasoning. 
A promising way to overcome these limitations is to empower LLMs with external tools, known as tool learning~\cite{qin2023tool,qu2024tool}. 
Tool learning not only enhances LLMs' performance on existing tasks but also allows them to tackle tasks that were previously beyond their reach.
Besides, the ability to use tools is a crucial hallmark on the path to advanced intelligence.

\begin{figure}[t]
    \centering
    \includegraphics[width=0.45\textwidth]{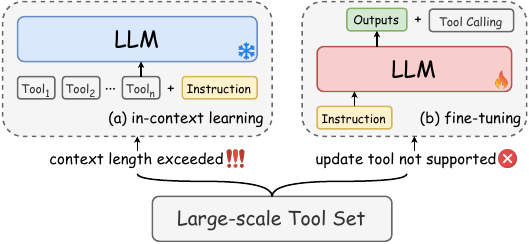}
    \caption{Illustration of two tool-learning approaches in LLMs: (a) in-context learning and (b) fine-tuning. The challenges posed by the extensive and frequently updated tools require the external tool retrieval component. }
    \label{fig:intro}
\vspace{-1em}
\end{figure}


Existing tool learning methods have preliminarily demonstrated that LLMs could effectively utilize specific tools to complete corresponding tasks.
They either leverage LLMs' in-context learning ability to facilitate tool usage with tool descriptions~\cite{NEURIPS2023_77c33e6a} or fine-tune LLMs to integrate tool learning capabilities into parameters, e.g., Toolformer~\cite{NEURIPS2023_d842425e}.
However, as illustrated in Figure~\ref{fig:intro}, existing methods still face significant challenges in real-world scenarios due to the following reasons. 1) The number of tools is usually vast, making it impossible for LLMs to handle them all with the limited input length of in-context learning. 2) Tools would frequently and irregularly update, rendering finetuning-based approaches costly and impractical.
Therefore, a tool retrieval component, which aims to select appropriate tools from a large-scale tool set, is essential for LLMs.

\begin{figure}[t]
    \centering
    \includegraphics[width=0.95\columnwidth]{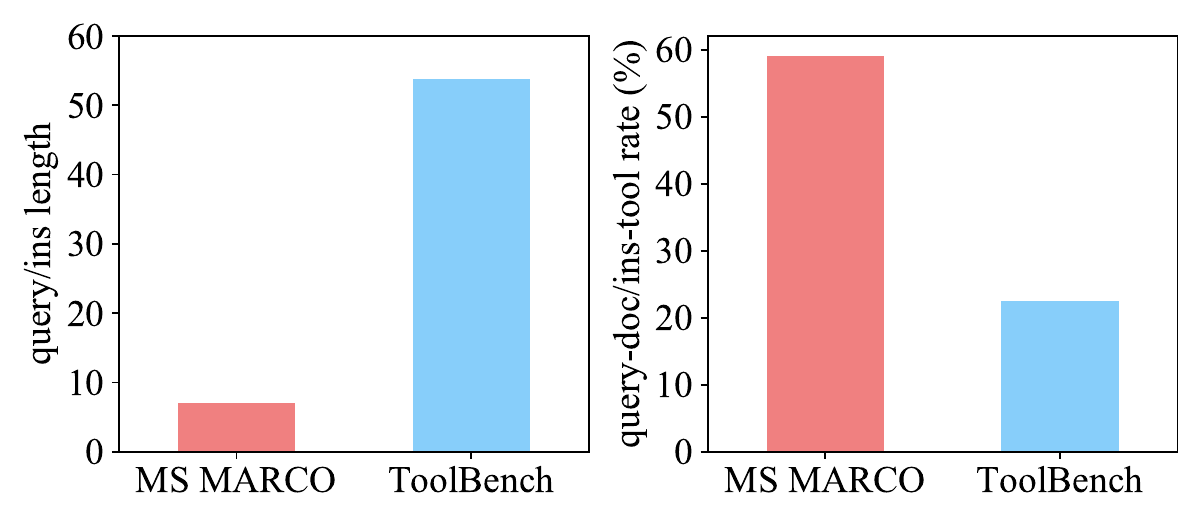}
    \caption{Comparison between the document retrieval and tool retrieval datasets. Tool retrieval presents more challenges due to the complex instructions (in the left figure) and the lower reputation rate (in the right figure). }
    \label{fig:intro_statistics}
    \vspace{-1em}
\end{figure}

Despite the practicality and necessity, tool retrieval has been inadequately studied. Some approaches have adopted traditional document retrieval methods to retrieve tools for LLMs~\cite{li-etal-2023-api,qin2023toolllm}. However, we argue that they overlook the unique challenges of tool retrieval for LLMs: 1) Complex usser instructions and tool descriptions. As illustrated in Figure~\ref{fig:intro_statistics}, compared with document retrieval, user instructions are usually ambiguous and complex, and the reputation rate between instructions and corresponding tool descriptions is much lower. Unfortunately, the retriever model is typically limited in its capacities because of the efficiency requirements, which makes tool retrieval more difficult and challenging.
2) Misalignment between tool retrieval and tool usage models. Previous approaches deploy the tool retriever separately from the downstream tool-usage model, which hinders the LLM from knowing which tools are really useful from the tool-usage perspective. Thus, it will result in a tool recognition gap between the tool retriever and tool usage model, degrading the tool-use performance further.

To address the above issues, we propose to enhance tool retrieval with iterative feedback. Our motivation is to utilize the LLM to enhance the comprehension ability of the tool retriever and bridge the gap between the two independent models.
At each iteration, we conduct a feedback generation process by asking the LLM to provide feedback step-by-step, conditioned on the user instruction and retrieved tools from the retriever.
The LLM will first comprehend the instruction and tool functionalities thoroughly, and then assess the effectiveness of those retrieved tools. According to the assessment, the LLM will refine the user instruction to improve the tool retrieval process. The refined instruction will substitute previous user instruction and be used to retrieve a new list of tools from the tool set.
In the next iteration, the new candidate tool list will be fed into the LLM for a new round of LLMs' feedback. During this iterative process, the tool retriever is expected to provide more appropriate tools for the tool-usage model. In this manner, the comprehension capability and tool preference of LLMs could be progressively incorporated into the retriever, and thus the tool retriever's performance could be continuously enhanced. We build a comprehensive tool retrieval benchmark, named TR-bench. The benchmark takes into account real-world practices with updated tools, and therefore encompasses both in-domain and out-of-domain settings. The experimental results show our approach achieves the best performance among the current methods with both in-domain and out-of-domain settings.

The key contributions are summarized:
\begin{itemize}
\item We identify the importance of tool retrieval in tool learning and present the distinct challenges of tool retrieval.
\item We propose to enhance tool retrieval with iterative feedback from the LLM. By leveraging iterative feedback, the tool retriever model gets continual improvements, ultimately reducing the misalignment between them.
\item We build a comprehensive tool retrieval benchmark with in-domain and out-of-domain settings, which will also aid future tool retrieval research. The extensive experiments demonstrate superior performance of our approach.
\end{itemize}


\section{Related Work}
\subsection{Tool Learning in LLMs}
Tool learning aims to equip LLMs with external tools to enhance and expand their capabilities~\cite{ruan2023tptu,wang2024mint,huang2024metatool}.
Generally, existing tool learning methods could be categorized into in-context learning and fine-tuning approaches. The former approach encourages LLMs to use tools with descriptions, documentation, or demonstrations~\cite{yuan2024easytool,du2024anytool,mu2024adaptive}, while the latter one trains the parameters of LLMs using specially created tool-use datasets~\cite{NEURIPS2023_8fd1a81c,tang2023toolalpaca,gao2024confucius}. However, no matter whether the in-context learning or fine-tuning approach encounters severe challenges in real-world scenarios, where the candidate tools are extensive and frequently updated. Therefore, it is crucial to equip LLMs with a tool retrieval component to select appropriate tools from a large-scale tool set. Recent works have proposed a stopgap measure through traditional document retrieval~\cite{patil2023gorilla,qin2023toolllm,zheng-etal-2024-toolrerank}, task decomposition~\cite{anantha2023protip,huang-etal-2024-planning} and graph-based methods~\cite{qu2024colt}. In this work, we aim to develop a method specialized for enhancing the tool retriever.

\subsection{Document Retrieval}
Early popular document retrieval methods rely on sparse retrieval that calculates the relevance of documents to a query based on the frequency of query terms in each document, e.g., BM25~\cite{10.1561/1500000019}.
With the development of language models \cite{devlin-etal-2019-bert}, 
the dense retrieval \cite{zhao2024dense,mitra2017neural} paradigm has gained considerable attention in the research community. 
By encoding queries and documents into high-dimensional vector representations and computing their relevance scores through inner product calculations, the paradigm can capture semantic relationships between queries and documents, thereby 
enhancing retrieval performance \cite{karpukhin-etal-2020-dense}. However, tool retrieval presents unique challenges, rendering traditional document retrieval methods suboptimal. We address these challenges by harnessing LLMs' feedback to iteratively refine the tool retrieval process.

\begin{figure*}[!t]
  \centerline{\includegraphics[width=2.0\columnwidth]{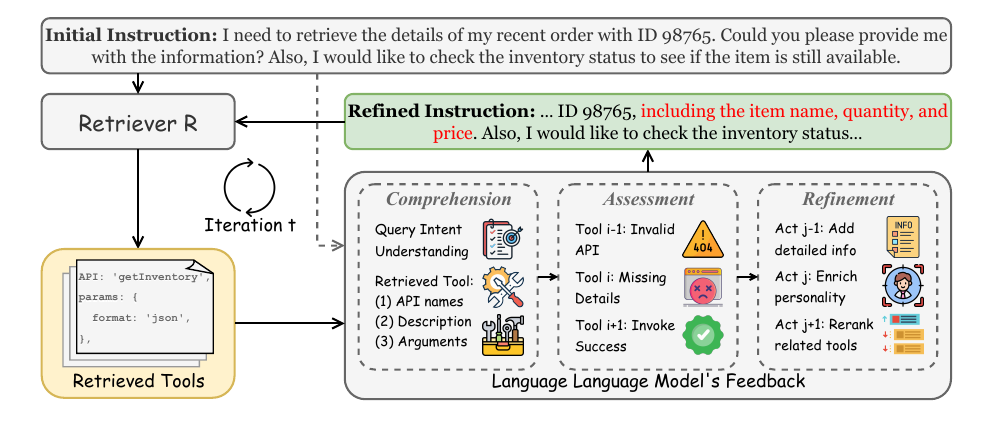}}
  \caption {Illustration of our proposed iterative tool retrieval method. At each iteration, the LLM follows a three-step feedback generation process, which includes comprehension, assessment, and refinement, to improve the instruction.}
  \label{MainFigure}
\end{figure*}

\section{Preliminaries}
\subsection{Task Definition}
Given a user's instruction, tool retrieval aims to select a small number of tools, which could aid the LLM in answering the instruction, from a large-scale tool set.
Formally, we define the user instruction as $q$ and the tool set as $D=\{d_1, d_2, ..., d_N\}$, where $d_i$ represents the description of each tool and $N$ is the total number of tools.
The retriever model $R$ needs to measure the relevance $R(q, d_i)$ 
between the instruction $q$ and each tool description $d_i$, and return $K$ tools, denoted as $D=\{d_1, d_2, ..., d_K\}$.

\subsection{Dense Retriever}
Dense retriever usually leverages the encoder-based LLM to encode the user instruction $q$ and a tool description $d$ into dense 
embeddings $E(q)$ and $E(d)$, respectively.
Then, it could measure the relevance between $q$ and $d$ by calculating the similarity score between these two embeddings, denoted as $R(q, d)=sim(E(q),E(d))$.

Dense retriever is trained via the contrast learning objective, which is designed to minimize the distance between the instruction embedding and embeddings of positive tools (the instruction's ground-truth tools) while maximizing the distance between the instruction embedding and embeddings of negative tools. 
The objective can be formulated as follows,
\begin{equation}
    \mathcal{L} = -\frac{1}{B}\sum\limits_{i=1}^{B}log\frac{e^{R(q_i,d_i^+)}}{e^{R(q_i,d_i^+)}+\sum_{j} e^{R(q_i,d_{ij}^-))}},
\end{equation}
where $B$ denotes the batch size, $d_i^+$ denotes the positive tool, and $d_{ij}^-$ represents the $j$-th negative tool to the instruction $q_i$.

However, due to the efficiency requirements, dense retrieval utilizes a dual-encoder architecture, which has limited ability to understand instructions. In this study, our goal is to improve the tool retrieval process with the feedback from the tool-usage model, i.e., the LLM.

\section{Methodology}
\subsection{Overview}
Recent studies have found that LLMs show a great capability in acting as a critic~\cite{NEURIPS2023_91f18a12} and could provide comprehensive feedback to improve performance across a range of tasks~\cite{NEURIPS2023_91edff07,asai2023self}. Inspired by those observations, we propose an innovative framework that leverages the LLM's feedback to improve the tool retrieval process iteratively. 
Different from approaches which focus on feedback from execution results after tool execution step~\cite{yao2023react,wang-etal-2024-llms-imaginarium}, we obtain LLMs' feedback before the actual tool execution step, i.e., right after the tool retrieval step.

As illustrated in Figure~\ref{MainFigure}, at each iteration, the LLM will provide feedback on the current-turn retrieval results. Specifically, the LLM will first comprehend the 
user instruction and tool functionalities thoroughly. Then, it will assess the effectiveness of those retrieved tools for handling the instruction.
Based on the assessment, the LLM could provide a
refinement to the retrieval model, refining the user instruction if necessary. To ensure that the retriever model is aware of the iteration round, we conduct an iteration-aware feedback training process to adapt the retriever model with continuously refined user instructions.

\subsection{Feedback Generation}
Assuming at the iteration step $t$, given the refined instruction $q^{t}$, we could utilize retriever model $R$ to retrieve a list of top-$K$ tools $\{d^{t}_1, ..., d^{t}_K\}$.
We then conduct a three-step feedback generation process by feeding those retrieved tools and associated tool descriptions into the LLM as follows.

\textbf{Comprehension}.
Firstly, the LLM is prompted to give comprehension on both the given instruction and retrieved tools. 
The prompt provided to LLM includes two parts: 
(1) summarize the abstract user goals by ignoring detailed entity information in the given instruction;
(2) understand the functionalities of retrieved tools, focusing on the category, name, description, input and output parameters of given tools.
This step can be formulated as, 
\begin{equation}
    F_C=LLM(P_C, q^{t}, \{d^{t}_1, ..., d^{t}_K\}),
\end{equation}
where $F_C$ denotes LLM's comprehension output and $P_C$ denotes the prompt provided to LLM.

\textbf{Assessment}.
The LLM will assess the effectiveness of retrieved tools for handling the instruction based on its comprehension of the user's itent and tool functionalities. 
The assessment is conducted from two perspectives: 1) identify which of the user's goals could and could not be solved by the retrieved tools with corresponding reasons;
and 2) analyze whether the ranked order of retrieved tools corresponds with their significance in addressing the user's intent with specific reasons.
The step can be formulated as, 
\begin{equation}
    F_A=LLM(P_A, q^{t}, \{d^{t}_1, ..., d^{t}_K\}, F_C),
\end{equation}
where $F_A$ denotes the LLM's assessment output.

\textbf{Refinement}.
Lastly, the LLM will refine user instruction based on its assessment.
Specifically, we ask the LLM to determine whether the refinement is necessary based on the two following questions: 1) Whether all the user's goals have been solved by currently retrieved tools, 2) and whether all existing appropriate tools are given the highest ranking priorities by the retriever.
If one of the answers is not ``yes'', we prompt the LLM to provide a potential refinement for retrieval improvement. 
Otherwise, the LLM will directly return a special token ``N/A'' without conducting any refinement.

The feedback from the LLM is finalized made on the current user instruction $q^t$.
Specifically, we prompt the LLM to generate refined instruction with enriched information in two dimensions: 1) more detailed and personalized content about those user's intent which have not been solved by current tools, helping the retriever explore other relevant tools;
(2) more
scenario-specific tool-usage information about existing appropriate tools, helping the retriever give higher ranking priority to those tools.
This step can be formulated as, 
\begin{equation}
    F_R=LLM(P_R, q^{t-1}, \{d^{t-1}_1, ..., d^{t-1}_K\}, F_A),
\end{equation}
where $P_R$ is the corresponding prompt and $F_R$ denotes LLM's refinement output, i.e., the new refined instruction $q^{t+1}$.

\subsection{Iteration-Aware Feedback Training}
We concatenate a special token ``Iteration $t$'' in front of the instruction, where $t$ is the instruction's iteration step (e.g., ``Iteration $t-1$'' for $q^{t-1}$  and ``Iteration $t$'' for $q^t$). 

We also employ the hard negative sampling in training.
Concretely, for each given instruction, we randomly sample an incorrect tool from the retrieved top-$K$ tool list.
The high similarity scores of those tools indicate that they are prone to be mistaken as correct tools by the retriever.
In feedback training, we utilize those tool-instruction pairs as hard negative samples.
Then the loss function for each iteration could be calculated as, 

{
\small
\begin{equation}
\begin{aligned} 
    \mathcal{L} = &\\
    -\frac{1}{B}&\sum\limits_{i=1}^{B}log\frac{e^{R(q_i,d_i^+)}}{e^{R(q_i,d_i^+)}\!+\!
    \sum_{j\neq i}e^{R(q_i,d_{ij}^-))}\!+\!
    \sum e^{M(q_i,d_{ij}^H)}}, 
\end{aligned}
\label{loss}
\end{equation}
}
where $d_{ij}^H$ denotes the hard negative sample.
By distinguishing the subtle differences in the tool descriptions, the retriever could achieve a deeper understanding of the tool functionalities and their relation with user instructions.

Then the final training objective could be formulated as the sum of losses in each iteration as follows,
\begin{equation}
\mathcal{L}_{feedback} = \sum_{t=1}^T \alpha^t\mathcal{L}(q^t),
\end{equation}
where $\alpha^t$ is a balancing factor and $L(q^t)$ is the loss function calculated by Equation~\ref{loss} based on the refined user instructions $q^t$ in the $t$th iteration.
In this way, the LLM's comprehensive knowledge of the user requirements could be injected into the retriever through those refined instructions.
Besides, with the aid of iteration-aware tokens and joint-training manner, the retriever could maintain a balance between newly learned knowledge and previously acquired knowledge.

\subsection{Inference}
At the time of inference, the feedback generation process keeps working while the feedback training process ceased. 
The retriever will update the candidate tool list based on the refined user instruction from LLM's feedback iteratively, until output the final retrieved tools.

Concretely, assume that we have obtained a retriever $R$ after the feedback training.
For each initial test instruction $q^0_{test}$, we add a special token ``Iteration 0'' in front of the instruction.
Then we use the trained retriever $R$ to retrieve an initial tool list $D^0_{test}$, containing $K$ candidate tools $\{d_1,d_2,...,d_K\}$. 
The retrieved $D^0_{test}$ and $q^0_{test}$ will be fed to the LLM for feedback generation, including instruction refinement, as discussed in Section 4.2.
After obtaining the refined instruction $q^1_{test}$, we add a token ``Iteration 1'' to it and then input it to $R$ for the next-round tool retrieval.
Then, we can get an updated tool list $D^1_{test}$ for a new round of feedback generation.
As such, we could obtain a final tool list $D^T_{test}$ after $T$ iterations.

\section{Experiments}
\subsection{Setup}
\textbf{Datasets and evaluation}. To assess the tool retrieval performance of models, we conduct an experiment on tool retrieval benchmark, referred to as \textbf{TR-bench}, based on three datasets, including ToolBench~\cite{qin2023toolllm}, T-Eval~\cite{chen2023t}, and UltraTools~\cite{huang2024planning}. To address real-world requirements, we conduct evaluations in both \textit{in-domain} and \textit{out-of-domain} settings. Specifically, the training set is from ToolBench, while the test set of ToolBench is employed for in-domain evaluation, and the test sets from T-Eval and UltraTools are used for out-of-domain evaluation.  The statistics of TR-bench are summarized in Table~\ref{dataset_statistics}.

\begin{table}[tbp]
\centering
 \scalebox{0.75}{
\begin{tabular}{cccc}
\toprule
&scenarios& \# instructions &\# tool set\\
\midrule
\multirow{4}{*}{\makecell[c]{Training \\Set}} &ToolBench-I1& 86,643& - \\
&ToolBench-I2&84,270& -\\
&ToolBench-I3 &25,044& -\\
&ToolBench-All &195,937&-\\
\midrule
\multirow{4}{*}{\makecell[c]{In-domain \\Evaluation}} &ToolBench-I1& 796& 10,439 \\
&ToolBench-I2&573& 13,142\\
&ToolBench-I3 &218& 1,605\\
&ToolBench-All &1,587&13,954\\
\midrule
\multirow{2}{*}{\makecell[c]{Out-of-domain \\Evaluation}} &T-Eval& 553& 50 \\
&UltraTools&1,000& 498\\
\bottomrule
\end{tabular}}
\caption{Statistics of the TR-bench, which is conducted from ToolBench~\cite{qin2023toolllm}, T-Eval~\cite{chen2023t}, and UltraTools~\cite{huang2024planning}. }
\label{dataset_statistics}
\end{table}

Following ToolBench, we adopt the Normalized Discounted Cumulative Gain (NDCG)~\cite{10.1145/582415.582418}, an ideal metric for tool retrieval to evaluate the quality of retrieved tools.
In our evaluation, we report NDCG@$m$ ($m=1,3,5,10$), calculated according to the position of each golden tool among top-$m$ candidates tools retrieved by the tool retriever. 
Thus, the more accurately the tool retriever can retrieve correct tools, the higher the NDCG@$m$ score will be.

\begin{table*}[t]
\centering
\small
\resizebox{\linewidth}{!}{
\begin{tabular}{@{}l|ccc|ccc|ccc|ccc@{}}
\toprule
\multirow{2}{*}{\textbf{Methods}} & \multicolumn{3}{c|}{\textbf{\textsc{Single-tool (I1)}}} & \multicolumn{3}{c|}{\textbf{\textsc{Category (I2)}}} & \multicolumn{3}{c|}{\textbf{\textsc{Collection (I3)}}} & \multicolumn{3}{c}{\textbf{\textsc{All}}} \\ \cmidrule(lr){2-13}
&N@1  &N@3  &N@5 &N@1  &N@3  &N@5  &N@1  &N@3  &N@5  &N@1  &N@3  &N@5  \\ \midrule
BM25 &$18.37$  &$17.97$  &$19.65$  &$11.97$  &$9.85$  &$10.95$  &$25.23$  &$18.95$  &$20.37$  &$15.84$  &$13.98$  &$15.63$  \\
Ada Embedding &$57.52$ &$54.90$  &$58.83$  &$36.82$  &$28.83$  &$30.68$  &$54.59$  &$42.55$  &$46.83$  &$46.59$  &$41.06$  &$43.95$  \\ 
ToolRetriever &$84.20$  &$89.59$  &$89.65$  &$68.24$  &$77.43$  &$77.90$  &$81.65$  &$87.24$  &$87.13$  &$75.73$  &$83.19$  &$83.06$  \\ 
\textbf{Ours} &\bf 90.70  &\bf 90.95  &\bf 92.47  &\bf 89.01  &\bf 85.46  &\bf 87.10  &\bf 91.74  &\bf 87.94  &\bf 90.20  &\bf 88.53  &\bf 87.00  &\bf 88.83  \\ 
\textbf{\% improve} &7.72\% & 1.52\% &3.15\%  &30.44\%   &10.37\%  &11.81\%   &12.36\%   &0.80\%    &3.52\%   &16.90\%    &4.58\%    &6.95\%  \\ 
\bottomrule
\end{tabular}}
\caption{In-domain evaluation on TR-bench in terms of NDCG@$m$ under scenarios including single-tool (\textit{I1}), intra-category multi-tool (\textit{I2}), intra-collection multi-tool (\textit{I3}), and the whole data (\textit{All}). \% improve represents the
relative improvement achieved by our method over the previously best tool retrieval method.}  
\label{main_results}
\end{table*}
\begin{table*}[t]
\centering
\small
\begin{tabular}{@{}l|cccc|cccc@{}}
\toprule
\multirow{2}{*}{\textbf{Methods}} & \multicolumn{4}{c|}{\textbf{\textsc{T-Eval}}} & \multicolumn{4}{c}{\textbf{\textsc{UltraTools}}} 
\\ \cmidrule(lr){2-9}
&N@1  &N@3  &N@5 &N@10 &N@1  &N@3  &N@5 &N@10 \\ \midrule
BM25 &$52.12$  &$43.19$  &$45.23$  &$52.91$  &$15.10$  &$14.13$  &$16.03$  &$18.34$  \\
Ada Embedding &$80.11$ &$69.11$  &$71.95$  &$79.62$  &$31.46$  &$33.75$  &$39.91$  &$46.40$  \\ 
ToolRetriever &$82.10$  &$72.03$  &$74.15$  &\bf 80.76  &$48.20$  &\bf 47.73 &$53.01$  &$58.93$  \\ 
\textbf{Ours} &\bf 84.45  &\bf 73.31  &\bf 74.45  &$80.25$  &\bf 49.30 &47.50  &\bf 54.30  &\bf 59.92 \\ 
\textbf{\% improve} & 2.86\% & 1.78\% & 0.40\% & -0.06\% &2.28\% &-0.48\%   &2.43\%   &1.68\% \\ 
\bottomrule
\end{tabular}
\caption{Out-of-domain evaluation on TR-bench in terms of NDCG@$m$ under two scenarios, T-Eval~\cite{chen2023t} and UltraTools~\cite{huang2024planning}. \% improve represents the
relative improvement achieved by our method over the previously best tool retrieval method.}  
\label{ood}
\end{table*}

\textbf{Baselines}.
We compare our method against representative retrieval methods.
1) BM25~\cite{10.1561/1500000019}: the classical sparse retrieval method; 
2) Ada Embedding: the closed-sourced OpenAI’s text-embedding-ada-002 model\footnote{\url{https://platform.openai.com/docs/guides/embeddings/embedding-models}.};
3) ToolRetriever~\cite{qin2023toolllm}: a dense retrieval approach specifically finetuned on tool retrieval datasets. 

\textbf{Implementation details}. We employ Sentence-BERT~\cite{reimers-gurevych-2019-sentence} to train our retriever model based on BERT-base~\cite{devlin-etal-2019-bert}. 
We set the learning rate to $2e{-5}$ with 500 warm-up steps. The batch size in training is set to $64$. We utilize ChatGPT (gpt-3.5-turbo-0125)\footnote{\url{https://openai.com/index/introducing-chatgpt-and-whisper-apis/}.} as the LLM for giving feedback.
The number of tool candidates $K$, the balancing factor $\alpha$, and the iteration round $T$ are set to 10, 1, and 3, respectively. We have trained the model several times to confirm that the improvement is not
a result of random chance and present the mid one. Our experiments were conducted on four NVIDIA
A6000 GPUs with 48 GB of memory.

\subsection{Main Results}
\textbf{In-domain evaluation}. The results of the in-domain evaluation are reported in Table~\ref{main_results}.
It is observed that non-finetuned retrieval methods, i.e., BM25 and Ada Embedding, perform much worse than other finetuned methods.
This is reasonable since non-finetuned methods have not been specifically adopted for tool retrieval.
While Tool Retriever outperforms non-finetuned methods, the performance is still not satisfying.
In comparison, our proposed method consistently outperforms all finetuned and non-finetuned baselines.
Significantly, our method maintains strong performance in the intra-category multi-tool (I2) scenario, even as other methods' performance declines, demonstrating the robustness of our proposed method across different scenarios.
The above results prove the effectiveness of our method in enhancing tool retrieval accuracy, particularly in challenging scenarios with multi-tools.

\textbf{Out-of-domain evaluation}. Since the tools are usually frequently updated in real-world, we further test all methods in the out-of-domain setting, where the training data from ToolBench and the test data from T-Eval and UltraTools are used.
The experimental results are shown in Table~\ref{ood}.
We could observe that our method significantly outperforms other baselines across both scenarios.
This demonstrates that our method not only excels in in-domain benchmarks but also maintains robust performance across varied scenarios, revealing its generalization ability of tool retrieval. 

We further compare the tool usage performance of our method with ToolRetriever in the I2 scenario. We adopt ToolLLaMA~\cite{qin2023toolllm} which is trained on LLM-annotated solution path as the tool usage model, and use “pass rate” and “win rate” as evaluation metrics. 
Our method achieves 75.6\% for pass rate compared to ToolRetriever's 68.5\%, and 65.9\% for win rate compared to ToolRetriever's 60.8\%. The results demonstrates the performance improvement in tool usage, benefiting the entire tool learning process.

\begin{table}[t]
\centering
\small
    \begin{tabular}{l|cccc}
    \toprule
    Methods & N@1 & N@3 & N@5 & N@10 \\
    \midrule
    Ours & 89.01 & 85.46 & 87.10 & 88.41 \\
    \midrule
    \textit{w/o warm-up} & 85.51 & 81.36 & 84.47 & 86.92 \\
    \textit{w/o hard-negative} & 86.04 & 80.41 & 84.00 & 85.98 \\
    \textit{w/o joint} & 85.38 & 81.55 & 83.79 & 86.20 \\
    \textit{w/o joint \& hard-neg} & 83.77 & 77.67 & 81.21 & 83.69 \\
    \bottomrule
    \end{tabular}
    \caption{Ablation study of our method under the intra-category multi-tool (\textit{I2}) scenario.
    } 
    \label{ablation}
\end{table}

\begin{table}[tb]
\small
\centering
    \begin{tabular}{c|cccc|c}
    \toprule
    Iteration & N@1 & N@3 & N@5 & N@10 & Efficiency \\
    \midrule
    1 & $85.69$ & $80.48$ & $83.94$ & $86.27$ & 6.12s \\
    2 & $87.78$ & $83.48$ & $86.31$ & $88.26$ & 8.59s \\
    \bf 3 &\bf 89.01 &\bf 85.46 &\bf 87.10 &\bf 88.41 & 10.30s \\
    \bottomrule
    \end{tabular}
    \caption{Analysis on iteration round under the intra-category multi-tool (\textit{I2}) scenario. The efficiency is measured by the time consumption to complete one user instruction.} 
    \label{iteration_round}
\end{table}

\begin{table}[t]
\small
\centering
    \begin{tabular}{c|ccc}
    \toprule
    Methods & N@1 & N@3 & N@5  \\
    \midrule
    ToolRetriever (BERT-based) & $68.24$ & $77.43$ & $77.90$  \\
    \bf Ours (BERT-based) &\bf 89.01 &\bf 85.46 &\bf 87.10 \\ \midrule
    ToolRetriever (RoBERTa-based) & $76.61$ & $69.81$ & $74.99$ \\
    \bf Ours (RoBERTa-based) &\bf 88.13 &\bf 85.41 &\bf 86.75 \\
    \bottomrule
    \end{tabular}
    \caption{Analysis on different base models under the intra-category multi-tool (\textit{I2}) scenario.} 
    \label{base_model}
\end{table}

\begin{table}[t]
\small
\centering
    \begin{tabular}{c|cccc}
    \toprule
    Embedding Size & N@1 & N@3 & N@5 & N@10  \\
    \midrule
    300 & $87.61$ & $83.49$ & $85.20$ & $86.50$ \\
    512 & $87.61$ & $82.85$ & $84.67$ & $85.81$ \\
    \bf 768 &\bf 89.01 &\bf 85.46 &\bf 87.10 &\bf 88.41  \\
    1024 & $88.66$ & $83.91$ & $85.94$ & $87.04$ \\
    2048 & $88.74$ & $83.95$ & $85.98$ & $87.43$  \\
    \bottomrule
    \end{tabular}
    \caption{Analysis on embedding sizes under the intra-category multi-tool (\textit{I2}) scenario.} 
    \label{embedding_size}
\end{table}

\subsection{Ablation Study}
We conduct ablation studies to investigate the efficacy of different components in our methods.
First, we remove the warm-up training by directly conducting our method on an retriever based on Sentence-BERT.
Then, we analyze the contribution of hard negative sampling in our method by removing the hard-to-distinguish samples from the training.
In addition, we assess the efficacy of joint training in our method, by substituting it with a loss $\mathcal{L}_{feedback} = \mathcal{L}(q^t)$, with respect to only the refined instructions $q^t$ at current iteration $t$.
Table~\ref{ablation} reports the ablation test performance (i.e., NDCG@$m$ ($m=1,3,5,10$)) under the intra-category multi-tool instructions (I2) scenario on ToolBench.

From the results, we can observe that our method achieves comparably high NDCG scores even without warm-up training, indicating that it does not heavily rely on prior tool-use knowledge.
When hard negative sampling is removed, the performance degradation illustrates that hard negative sampling could enable the model to discriminate between similar tool functionalities. 
Besides, the model's performance further declines when joint training is removed, demonstrating that the model could balance new and previous knowledge in this joint-training manner.

\begin{figure*}[t]
  \centerline{\includegraphics[width=1.0\textwidth]{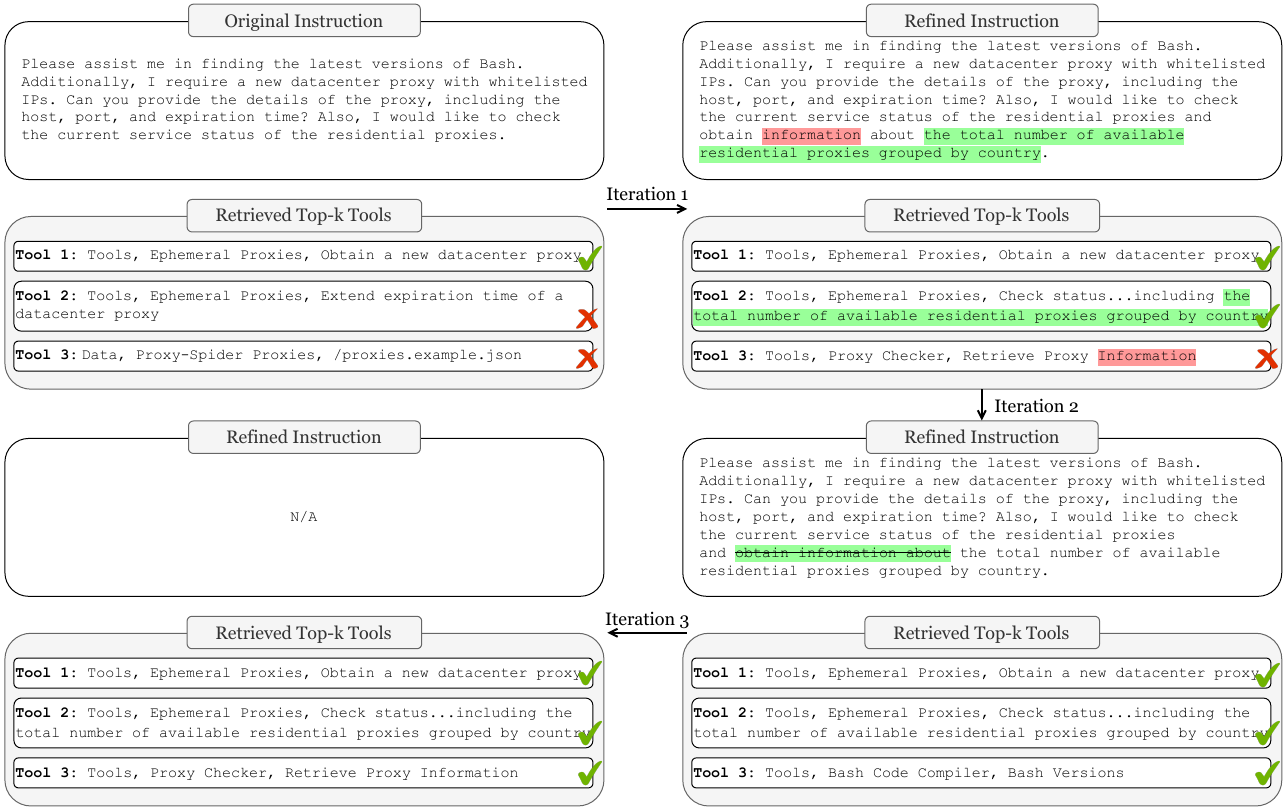}}
  \caption {Case study on the effect of user instruction refinement through 3 iterations. The original instruction is revised step-by-step,  leading to improved retrieval results.}
  \label{case_study}
\end{figure*}

\subsection{In-depth Analysis}
\textbf{Analysis on iteration round}. The iteration round is an important factor in our method. We conduct experiments to investigate changes in effectiveness and efficiency with different iteration round $T$. The results are presented in
Table~\ref{iteration_round}, and the efficiency is measured by the cost of time to complete one user instruction on average.

By analyzing the results in Table~\ref{iteration_round}, we gain two findings. 1) We could observe a continuous improvement as the iteration round increases.
This shows that the tool retriever progressively enhances its performance with the aid of LLMs' feedback. 2) In terms of time efficiency, we find that adding one additional round of refinement takes an average of 6.12s/instruction, primarily resulting from the time waiting for LLM's feedback when calling the OpenAI API.
As the number of iterations increases, we can see that the extra inference time required for each instruction decreases. 
This is due to the fact that there will be fewer instructions requiring refinement as retrieval performance improves.

\textbf{Analysis on base models}.
We further analyze the impact of different base models on the performance. Specifically, we replace the base model BERT in our method with another classic language model, RoBERTa~\cite{liu2019roberta}. The results are shown in Table~\ref{base_model}. As we can see, our method still achieves significant improvement over the baseline with the same RoBERTa model. Another observation is that RoBERTa is more effective in serving as a base model for the retrieval application, which benefits from its effective training strategies.  The improvements demonstrate the robustness of our method with different base models.

\textbf{Analysis on embedding sizes}.
Since the retriever model $R$ encodes the textual instruction and tool description into dense vectors, we explore the impact of the embedding size on retrieval performance. as shown in Table~\ref{embedding_size}. From the table, we can find that larger embedding sizes result in greater performance improvements compared to smaller embedding sizes. This is probably due to the fact that embeddings with larger sizes could accommodate more knowledge. However, when the embedding size increases from 768 to 2048, there is a slight decrease in performance. This suggests that a specific embedding size is sufficient, and larger embedding sizes may pose challenges to training. It is worth noting that larger embedding sizes necessitate higher training costs and increased inference memory. Therefore, we recommend an optimal embedding size of 768.

\subsection{Case Study}
As shown in Figure~\ref{case_study}, we conduct case study by using an example of instruction refinement to take a closer look at the effect of our method.

In the $1st$ iteration, we can observe that the refined instruction has included more detailed information (i.e., ``total number'') about the user's requirements than the original instruction, enabling the retriever to identify more appropriate tools (e.g., Check residential proxies service status).
This reveals that the comprehension capabilities of LLMs could be instilled into the retrieval process through feedback.
In the $2nd$ iteration, our method further refines the instruction by omitting irrelevant content (i.e., ``information'') which may mislead the retriever into retrieving incorrect tools (e.g., Retrieve Proxy Information). 
Another benefit of the refinement is that some correct tools (e.g., Bash Code Compiler) will move up in positions of the top-$K$ rankings, improving the overall retrieval performance.
In the $3rd$ iteration, our method showcases great decision-aware capabilities, where the iterative process could be terminated if no further refinement is deemed necessary.

\section{Conclusion and Future Work}
In this study, we concentrate on the crucial tool retrieval in the tool learning of LLMs. We have identified the bottleneck in the tool retrieval-usage pipeline as the limited tool retrieval model. We propose the unique challenges of the tool retrieval compared with document retrieval. To improve the current tool retrieval process, we propose leveraging the LLM's feedback to assess the retrieval results and provide detailed suggestions for refining user instructions. In order to integrate the retriever model into this iterative process, we implement iteration-aware feedback training. This will improve the tool retriever's capabilities and close the gap between tool retrieval and usage models. We conduct the TR-benchmark to comprehensively evaluate the models' ability in real-world tool retrieval scenarios. Our method demonstrates the best performance in both in-domain and out-of-domain settings.

In the future, we aim to improve this work from the following aspects. 1) Limited by the training speed, we have applied the offline feedback generation, where feedback is generated before training the tool retriever. We will also assess whether online feedback generation yields further improvements in the future. 2) Furthermore, as the tool retriever serves the subsequent tool usage model in tool learning, we intend to conduct further evaluations of the tool retriever models based on the subsequent tool usage results.

\section*{Limitations}
1) Undoubtedly, our iterative refinement will reduce the inference speed of the tool retrieval. The efficiency issue is inherent in approaches involving LLMs' interaction. We have evaluated the efficiency as the number of iterative rounds increases. Fortunately, we observed that the retrieval model can achieve a significant performance improvement after just a single round of LLMs' feedback compared to without feedback.
Furthermore, the performance enhancement of the tool retrieval is crucial for the subsequent tool usage model, ensuring that the correct tools are retrieved and lays the foundation for all subsequent steps of tool usage. Therefore, we believe that performance improvement is worthwhile despite some efficiency loss. We will also pay more attention to this issue in the future.
2) Similar to document retrieval, the used datasets in our work also contain ``false negative'' samples. For instance, some tools may be capable of handling the user's instruction but are not labeled as positive. This can disrupt the training and evaluation of tool retrieval and is a common limitation in many retrieval scenarios.
\section*{Ethics Statement}
The datasets used in our experiment are publicly released and labeled through interaction with humans in English. In this process, user privacy is protected, and no personal information is contained in the dataset. The scientific artifacts that we used are available for research with permissive licenses. And the use of these artifacts in this paper is consistent with their intended use. Therefore, we believe that our research work meets the ethics of the conference. 

\bibliography{acl_latex}

\end{document}